\documentclass{mva_style}
\usepackage{graphicx}
\usepackage{amsmath,amssymb} % define this before the line numbering.
\usepackage{color}
\usepackage{pifont}
\usepackage{bm}
\usepackage{subfig}
\usepackage{booktabs}
\usepackage{multirow}
\usepackage{mathtools}
\DeclareMathOperator*{\argmin}{arg\,min}

\finalcopy %Uncomment this line for the Camera-Ready Manuscript

\begin{document}
\title{Can fully convolutional networks perform well for general image restoration problems?}

\author{
  Subhajit Chaudhury\\
  Takatsu, Kawasaki City\\
   Kanagawa Prefecture, Japan\\
  {\tt subhajit.ju4u@gmail.com}\\
  \and
  Hiya Roy\\
  The University of Tokyo\\
  JAXA, Sagamihara, Japan\\
  {\tt hiya.roy@ac.jaxa.jp}\\
}

\maketitle

\section*{\centering Abstract}
\vspace*{-0.3cm}
\textit{
  We present a fully convolutional network(FCN) based approach for color image restoration. FCNs have recently shown remarkable performance for high-level vision problem like semantic segmentation. In this paper, we investigate if FCN models can show promising performance for low-level problems like image restoration as well. We propose a fully convolutional model, that learns a direct end-to-end mapping between the corrupted images as input and the desired clean images as output. Our proposed method takes inspiration from domain transformation techniques but presents a data-driven task specific approach where filters for novel basis projection, task dependent coefficient alterations, and image reconstruction are represented as convolutional networks. Experimental results show that our FCN model outperforms traditional sparse coding based methods and demonstrates competitive performance compared to the state-of-the-art methods for image denoising. We further show that our proposed model can solve the difficult problem of blind image inpainting and can produce reconstructed images of impressive visual quality.}

\vspace*{-0.2cm}
\section{Introduction}
\vspace*{-0.35cm}

Image restoration is the technique to convert a noisy image into a clean, original one. Common image restoration problems include image denoising and image inpainting. Image denoising is the method of removing the external noise (usually modeled as additive white Gaussian noise) to obtain the original uncorrupted image. Another form of corruption for image signal occurs in the form of missing pixel values. Image inpainting is used for predicting such missing pixel values or removing sophisticated patterns like superimposed texts from images and preserve the original image information. In this paper, we focus on the problems of image denoising and blind image inpainting.

Prominent techniques in image denoising perform modifications in the image domain itself. Notable methods in this category include total variation based image denoising \cite{Rudin:1992:NTV:142273.142312}, denoising by learning global image priors\cite{1467533} etc. Additionally, sparse coding-based image denoising is shown to produce an impressive performance which can also be extended to solve other image restoration tasks. Carefully engineered algorithms such as BM3D\cite{4271520} and its color variant CBM3D\cite{4378954}, which exploit similarity in appearance of different patches constitute the current state-of-the-art in image denoising.

Image inpainting can be broadly classified into two categories: non-blind inpainting and blind inpainting. While in non-blind inpainting, the algorithm is provided the prior knowledge of the spatial locations of the image with missing pixels or superimposed patterns that need to be restored, blind inpainting methods aim to solve a much more challenging problem of simultaneously identifying and restoring the corrupted pixels. In the field of non-blind image inpainting, region filling method\cite{1323101}, the sparse coding-based K-SVD\cite{4392496} model etc. have been proposed, however blind inpainting is a less mature field of study with limited implementations. To the best of our knowledge, SSDA based blind inpainting\cite{NIPS2012_4686} is the most notable work in blind image inpainting.

Our proposed method is inspired from classical domain transformation methods, where the image domain signal is converted to a new representation\cite{4099398} and coefficients are altered in the transformed domain to finally reconstruct the clean image. The proposed method is application specific and fully data-driven with no requirements of human designed filters which is the reason for superior restoration performance. We use a similar idea to that of Dong et al. \cite{7115171} for deep convolutional networks based image super-resolution and extend it to show that similar architectures can be used for image denoising and blind image inpainting, which is one of our major contributions in this paper. Moreover, our proposed solution is very simple to implement and consists of only convolutional layer (no pooling), which enables easy hardware implementation with fast image restoration performance. 

Autoencoders based image restoration techniques(like SSDA\cite{NIPS2012_4686}), compress the input image patch to a low-dimensional representation before decoding it to produce the final image reconstruction, which might lead to loss of information causing poor image restoration performance. In contrast, we maintain equal hidden unit dimension to the input image size throughout the network and perform the intermediate operations by filtering using convolution kernels. Since our proposed fully convolutional network does not compress input data, we believe that it is possible to perform better image restoration using the proposed model.
%We perform supervised training on the proposed FCN architecture by presenting our network with noisy(the corrupted image) and clean(desired image after removing the noise) image pairs. 
%In recent years, CNN has gathered massive popularity because of its outstanding performance on various challenging visual classification tasks\cite{NIPS2012_4824} due to the availability of huge visual datasets\cite{imagenet_cvpr09} and powerful GPUs for training. We also make use of these recent progress for efficiently training deeper networks for color image restoration.
Results in image denoising demonstrate that the proposed method is competitive with the state of the art methods. For image inpainting, although our model performs a much more difficult task of blind restoration, it demonstrates comparable visual reconstruction quality at par with non-blind inpainting methods. The capability of our model for blind inpainting of complex superimposed patterns is also a major contribution of this paper. 

\vspace*{-0.2cm}
\section{Proposed Method}
\label{sect:problem}
\vspace*{-0.35cm}
%In this section we describe the network architecture of the proposed FCN model along with the details on 
%the training procedure adopted for the network.

We map noisy images at the input to their corresponding clean image version by image domain transformation method. This mapping conceptually consists of three operations- (1) Basis projection i.e. projecting image patches onto learn dictionaries which is a novel representation of noisy images, (2) non-linear transformation for mapping the coefficients onto a new domain for representations of clean images and (3) reconstruction of clean image using weighted averaging on overlapping patches. Although the proposed concept is similar to image denoising using domain transformation, our method benefits from the unique feature of the ability to learn from data in an end-to-end fashion. Similar to the image super-resolution model presented in Dong et al. \cite{7115171}, we find that these three operations are similar to multidimensional filtering operations and can be performed by convolution operations. Hence, the mapping described above can be represented as a fully convolutional network.

\vspace*{-0.1cm}
\subsection{Model description}
\begin{figure}[h!]
	\captionsetup[subfloat]{labelformat=empty}
	\captionsetup{justification=centering}
	\centering
	{%
		\includegraphics[height=2.8cm,width=8cm]{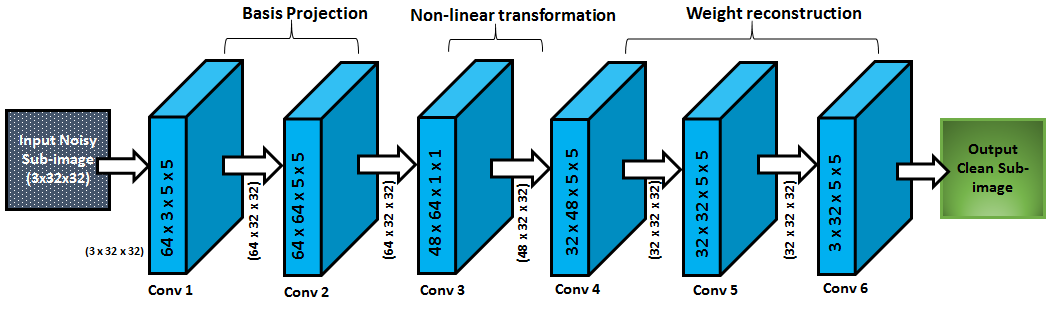}} \;
	
	\caption{Proposed Image restoration convolutional neural network}
	\label{fig:ircnn}
	
	\end{figure}

\vspace*{-0.4cm}
\textbf{\textit{IRCNN}(5-5-1-5-5-5)}: For solving the tasks of image denoising and blind image inpainting, we propose a  6 layered Image restoration CNN model (IRCNN) consisting of only convolution layers. Figure \ref{fig:ircnn} shows details on filter weights and number of convolution parameters for each layer. First two convolution layers of filter size $5\times5$ perform basic projection, next convolution layer ( $1 \times 1$ filter size) performs pixel-wise co-efficient alterations, and finally, last three convolution layers are responsible for converting the clean image representations to clean image.

\vspace*{-0.4cm}
\subsection{Training}
\label{sec:train}
\vspace*{-0.3cm}
%Our CNN model learns end-to-end mapping between the input noisy images and the output clean images. 
We optimize the network parameters $\Theta=\{W_i,B_i\},\;\;\; i=\{1,2,...,l\}$ by minimizing the loss between the set of clean images $\{\bm{Y}_i\}$ and images predicted $\{\hat{\bm{Y}}_i\}$ from the noisy image set $\{\bm{X}_i\}$. Let us define this overall mapping as $\hat{\vec{Y}}_i=F(\vec{X_i},\Theta)$. Then the optimal parameters are obtained as,
\vspace*{-0.3cm}
\begin{equation}
\hat{\Theta}=\argmin_{\Theta} \frac{1}{n}\smashoperator[r] {\sum_{i=1}^{n}} \|F(\bm{X}_i,\Theta)-\bm{Y}_i\|_2^2
\end{equation}
\vspace*{-0.3cm}

where $n$ is the number of images used for training the network. Minimizing the mean squared error between the clean and predicted image is performed by randomly sampling some smaller images from the clean/noisy images. Some pre-processing is done on these "sub-images" in the form of mean subtraction and normalization. In order for the size of the input and output sub-image to be same, we perform padded convolution in each layer. In our implementation, we used $3 \times 32 \times 32$ sized sub-images. For each kind of noise we produce the noisy image from the clean image and sample the same spatial location on each of these image pairs to produce a clean/noisy sub-image pairs.Training is done following standard mini-batch gradient descent approach(batch-size=256) with momentum update. 

%We implemented our model using Theano\cite{bergstra+al:2010-scipy,Bastien-Theano-2012} library in Python.

\begin{figure}[h!]
	\captionsetup[subfloat]{labelformat=empty}
	\captionsetup{justification=centering}
	\centering
	\subfloat["198054"]{%
		\includegraphics[height=2.3cm,width=1.6cm]{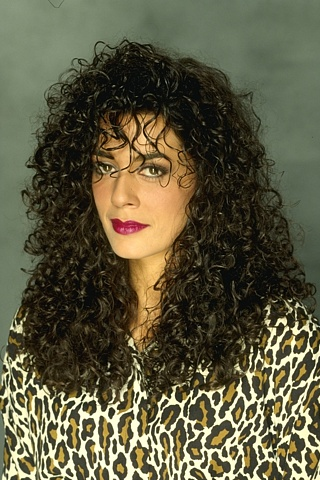}} \;
	\subfloat[Noisy, $\sigma=25$ ]{%
		\includegraphics[height=2.3cm,width=1.6cm]{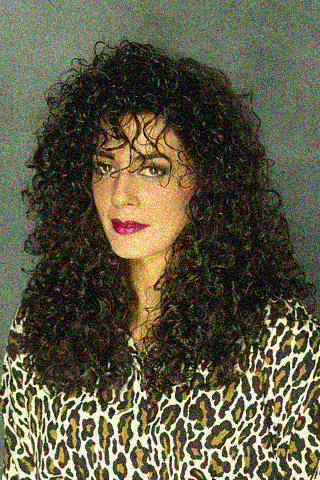}} \;
	\subfloat[CBM3D, 28.85dB]{%
		\includegraphics[height=2.3cm,width=1.6cm]{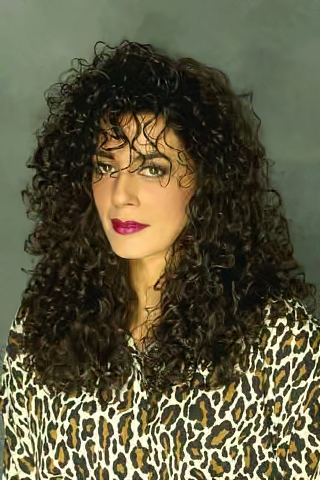}}  \;
	\subfloat[IRCNN, \textbf{29.56}dB]{%
		\includegraphics[height=2.3cm,width=1.6cm]{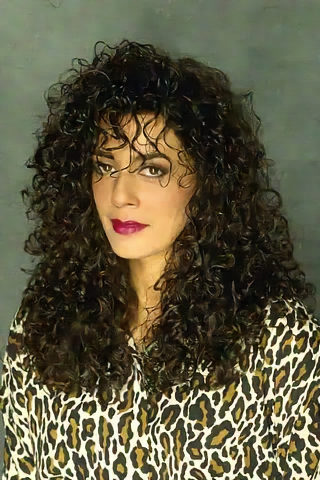}}\\

	\subfloat["Castle"]{%
		\includegraphics[height=2.3cm,width=1.6cm]{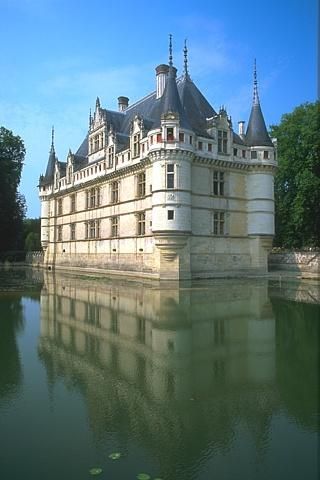}} \;
	\subfloat[Noisy, $\sigma=25$ ]{%
		\includegraphics[height=2.3cm,width=1.6cm]{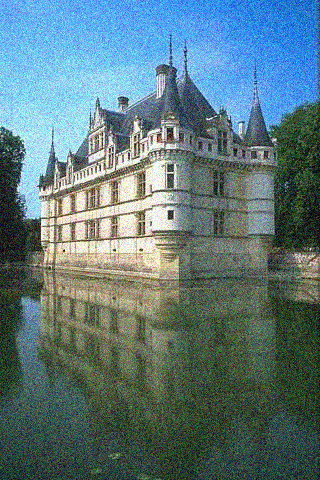}} \;
	\subfloat[CBM3D, \textbf{32.24}dB]{%
		\includegraphics[height=2.3cm,width=1.6cm]{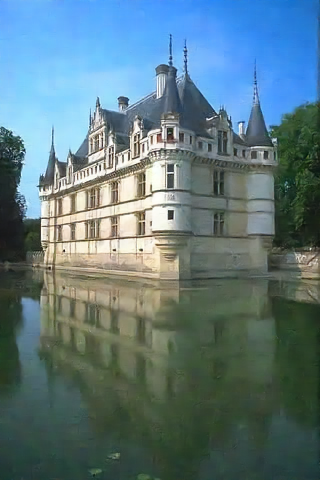}} \;
	\subfloat[IRCNN, 32.17dB]{%
		\includegraphics[height=2.3cm,width=1.6cm]{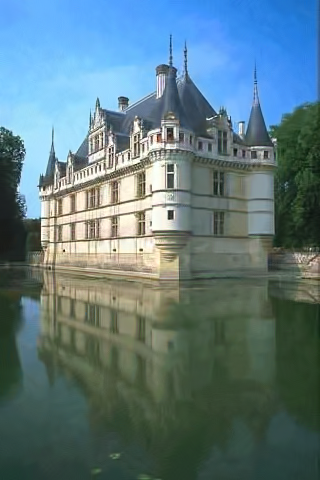}}\\
	
	\caption{Image denoising results(PSNR) on Berkeley segmentation dataset
	}\label{fig:berkley}
\end{figure}

\section{Experimental Results}
\label{sec:expres}
\vspace*{-0.4cm}
\subsection{Image denoising}
\label{sec:denoise}
\vspace*{-0.4cm}
Noisy images are created by corrupting clean images with additive white Gaussian noise. For our experiments we trained our network IRCNN for noise levels of $\sigma=25$ and $\sigma=50$. For training we extract sub-image pairs from original clean/noisy image pairs and train our network on these sub-image pairs.

For training our network we use data from two datasets:(1) Image-Net\cite{imagenet_cvpr09} (2) MSCOCO \cite{Lin2014}. We randomly choose 6000 images from each of the two datasets and corrupt each image with additive white Gaussian noise. From each such image pair, we choose 16 random samples of size $3 \times 32 \times 32 $, giving a total of 192,000 sub-image pairs. It took 4 days to train the network on a modern GPU, during which time 4000 passes over all the 192,000 sub-images were performed for IRCNN network. However, for testing purpose we used two test datasets (1)Berkeley segmentation dataset\cite{MartinFTM01} and (2)Pascal VOC 2012\cite{Everingham15} for evaluating our performance. Testing is performed by sliding window technique and averaging overlapping reconstructions. 
\begin{table*}[]
	\centering
	\caption{Image denoising performance for Berkeley segmentation dataset images}
	\label{my-label}
	\begin{tabular}{|c|ccc|cc|}
		\hline
		\multirow{2}{*}{Image} & \multicolumn{3}{c|}{$\sigma=25$}            & \multicolumn{2}{c|}{$\sigma=50$}    \\ \cline{2-6} 
		& KSVD  & CBM3D            & IRCNN         & CBM3D            & IRCNN          \\ \hline
		Castle                 & 31.19 & \textbf{32.24} & 32.17            & \textbf{28.67} & 28.66            \\
		Mushroom               & 30.26 & \textbf{31.20} & 30.92            & \textbf{27.77} & 27.60            \\
		Horse                  & 29.81 & 30.67            & \textbf{30.83} & 27.59            & \textbf{27.84} \\
		Kangaroo               & 28.39 & 29.19            & \textbf{29.30} & 26.37            & \textbf{26.45} \\
		Train                  & 28.16 & 28.72            & \textbf{28.88} & 24.52            & \textbf{25.06} \\ \hline
		Average                & 29.56 & 30.40            & \textbf{30.42} & 26.98            & \textbf{27.12} \\ \hline
	\end{tabular}
\end{table*}
\vspace*{-00cm}

Images from the Berkeley segmentation dataset, used in \cite{4392496}, were used to compare the performance of IRCNN with baseline method K-SVD\cite{4392496} and CBM3D\cite{4378954}, a state-of-the-art color image denoising method. For each image, experiments were performed $10$ times and the average PSNR value was reported. We used PSNR values reported by the authors in \cite{4392496} for the comparison. For CBM3D, we used the Matlab code provided by the authors for our evaluations. Table 1 shows the comparison of performance for image denoising with $\sigma=25$ and $\sigma=50$. On this small testing dataset, IRCNN produces a superior performance for 3 out of 5 images(for both $\sigma=25$ and $\sigma=50$) and has the best overall performance out-performing both sparse coding-based KSVD method and CBM3D method. We also tested with Convolutional Autoencoders 
%of size 5(32 filters)-pool(2)-5(16 filter)-pool(2)-5(16 filter)-upscale(2)-5(32 filters)-upscale(2)-5(3 filters) 
on 96000 image patches for 1000 epochs. The average PSNR for denoising task on the 5 images in Table 1 are 27.36dB and 25.06db for sigma=25 and 50 respectively. Since it is our own implementation and we believe that these CNN autoencoder results can be slightly improved by hyper-parameter tuning and more training with larger datasets, we do not report it in Table 1. Figure \ref{fig:berkley} show the qualitative comparison for image denoising.

\begin{figure}[]
	\centering
	\captionsetup{justification=centering}
	\subfloat[]{%
		\includegraphics[height=2.4cm,width=4cm]{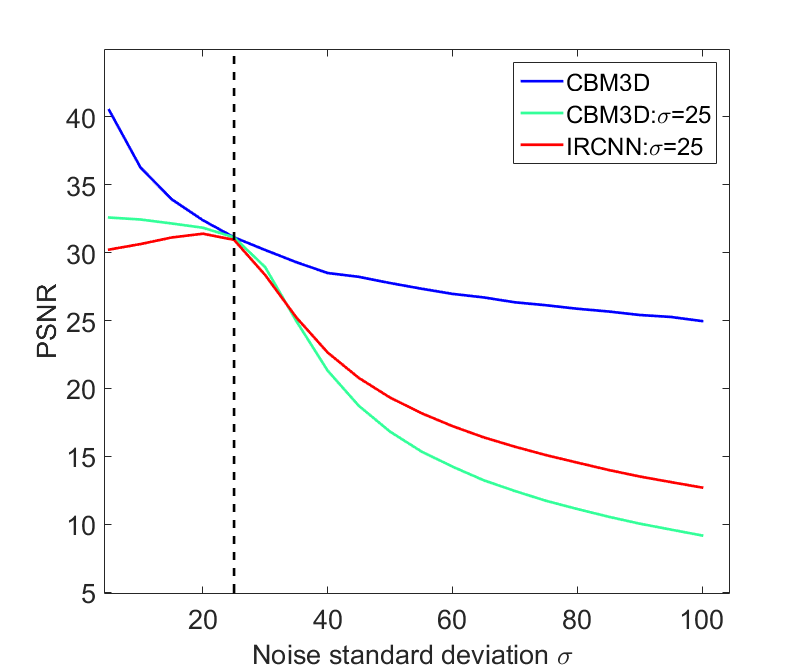}} \;\;
	\subfloat[]{%
		\includegraphics[height=2.4cm,width=4cm]{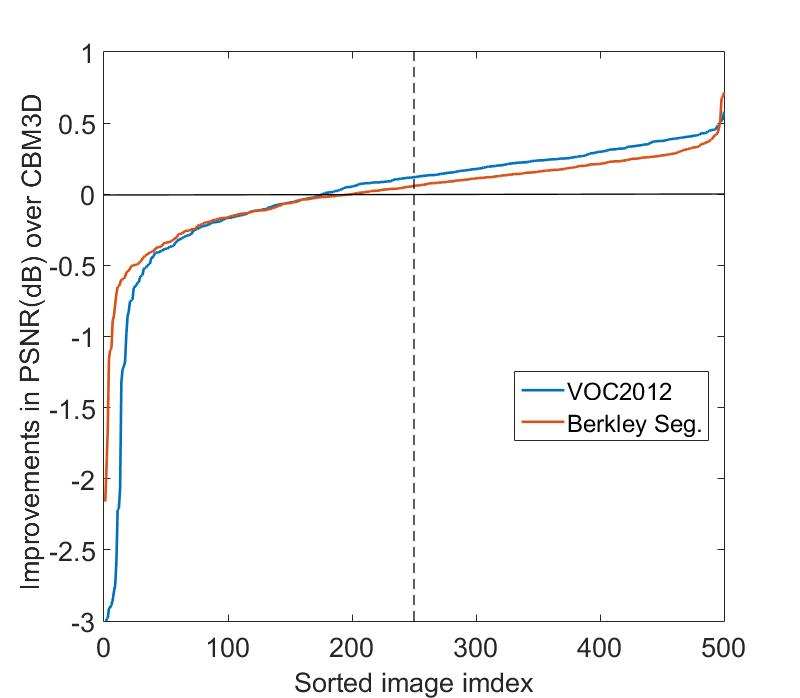}} \\
	
	\caption{(a) IRCNN trained at $\sigma=25$ vs CBM3D. (b) Improvements of IRCNN compared to CBM3D.
	}
	\label{fig:compare}
\end{figure}

For a more comprehensive comparison with CBM3D method, we tested the performance of both the methods on two large datasets of 500 images from Berkeley segmentation dataset\cite{MartinFTM01} and Pascal VOC 2012\cite{Everingham15} dataset. For each image in the dataset, experiments were performed 5 times and the average value was used. Improvements in PSNR achieved by our method, compared to CBM3D for $\sigma=25$ on both datasets is shown in figure \ref{fig:compare}(b). The comparisons between the CBM3D and IRCNN is shown in Table \ref{my-label}. These quantitative results demonstrate that the proposed IRCNN model performs at par with(often better than) state-of-the-art denoising methods.

We also test the IRCNN model trained at $\sigma=25$ for various other noise levels and plot the PSNR performance. The plot at various noise levels for the image "mushroom" from Berkeley segmentation dataset is shown in figure \ref{fig:compare}(a) which shows that our learned model produces competitive performance compared to CMB3D at $\sigma=25$ but performance deteriorates for other noise levels. To compare with similar effects in CBM3D, we fixed the input parameter to $\sigma=25$ for CBM3D. Similar performance is seen for CBM3D algorithm with knowledge of $\sigma =25$, although our learned network performs slightly better at higher noise levels. CBM3D provided with correct noise information produces a superior performance which is understandable. 

\vspace*{-0.2cm}
\subsection{Blind image inpainting}
\label{sec:inpaint}
\vspace*{-0.3cm}
We perform image inpainting task for images corrupted with (1) uniformly distributed missing pixels (2) complicated patterns like text. The training data for blind inpainting is same as that for image denoising. We make no attempt to change the network architecture for this task and perform training on IRCNN.
\vspace*{-0.2cm}
\subsubsection{Missing pixel inpainting}
\label{sec:miss}
\vspace*{-0.25cm}
Noisy images were created by randomly assigning $80\%$ of the pixel values in each channel as zeros and then 192,000 sub-images(similar to denoise case) were created by randomly sampling 16 images from each clean/noisy image pair. The training procedure is similar to the image denoising case.

\begin{figure}[h]
	\captionsetup[subfloat]{labelformat=empty}
	\captionsetup{justification=centering}
	\centering
	\subfloat[Image:castle]{%
		\includegraphics[height=2.5cm,width=2cm]{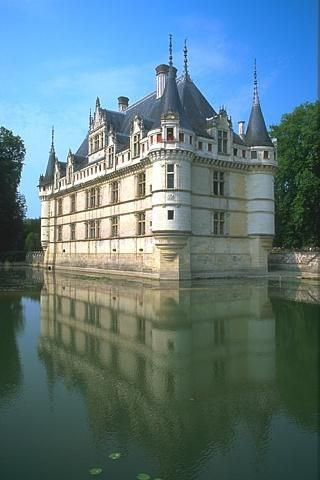}} \;\;
	\subfloat[Noisy, PSNR=6.68dB]{%
		\includegraphics[height=2.5cm,width=2cm]{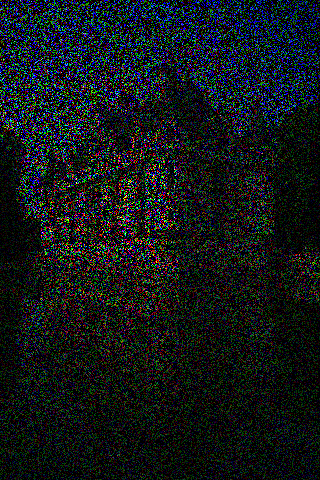}} \;\;
	\subfloat[Reconstructed, PSNR=28.74dB]{%
		\includegraphics[height=2.5cm,width=2cm]{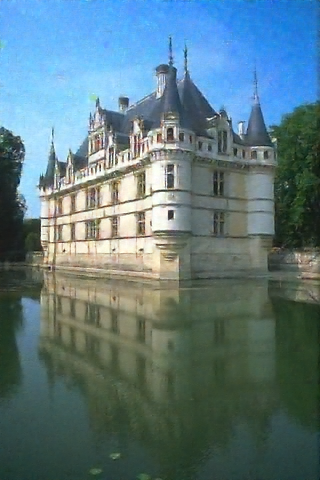}}\\
	
	\subfloat[Image:relativity]{%
		\includegraphics[height=2.5cm,width=2cm]{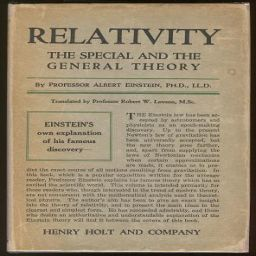}} \;\;
	\subfloat[Noisy, PSNR=5.89dB]{%
		\includegraphics[height=2.5cm,width=2cm]{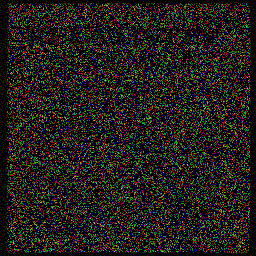}} \;\;
	\subfloat[Reconstructed, PSNR=28.11dB]{%
		\includegraphics[height=2.5cm,width=2cm]{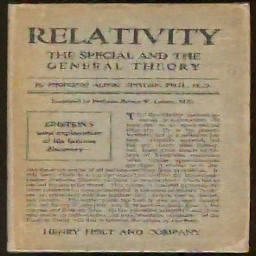}}\\
	\vspace*{-0.2cm}
	\caption{Missing pixel inpainting results on various images by IRCNN
	}\label{fig:miss}
\end{figure}

For $80\%$ missing pixel case we obtain a PSNR performance of 28.74dB for the image "castle" from Berkeley segmentation dataset. The best reconstruction performance of 29.65dB reported in \cite{4392496} by non-blind K-SVD inpainting technique. Our model has a lower PSNR performance compared to K-SVD because we solve a more difficult task of blind inpainting where the location of the missing pixels are unknown compared to the non-blind case where the information about the location of the missing pixel simplifies the inpainting problem to a large extent. 

Qualitatively our model shows good reconstruction quality, as seen from the results in figure \ref{fig:miss}. For the castle image, the reconstructed image is visually similar to the original clean image. For the "relativity" image, we observe that, while the text in the noisy image is not at all clearly visible, the image predicted by our model successfully restores readability for moderately large text. These qualitative and quantitative results demonstrate the effectiveness of our model for missing pixel restoration.

\subsubsection{Text removal}
\label{sec:text}
\vspace*{-0.3cm}
For text removal problem, noisy images were created by superimposing random texts on the clean images from 15 different font styles and font-size varying from 15pix to 25pix. Following similar methodology as previous methods, we create 192,000 sub-images by randomly sampling 16 images from each clean/noisy pair and training is done following standard mini-batch gradient descent with similar parameters as mentioned for previous tasks. Interestingly, we observed our model does not differentiate between the various tasks(denoising or inpainting) it is learning and takes almost similar time for learning the direct mapping between input and output in each case.
\vspace*{-0.19cm}
\begin{figure}[h!]
	\captionsetup[subfloat]{labelformat=empty}
	\captionsetup{justification=centering}
	\centering
	\subfloat[Clean Image]{%
		\includegraphics[height=2cm,width=3.5cm]{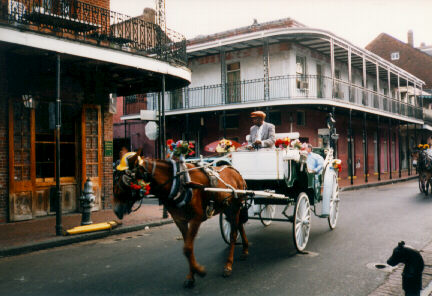}} \;
	\subfloat[Corrupted image, PSNR=15.05dB]{%
		\includegraphics[height=2cm,width=3.5cm]{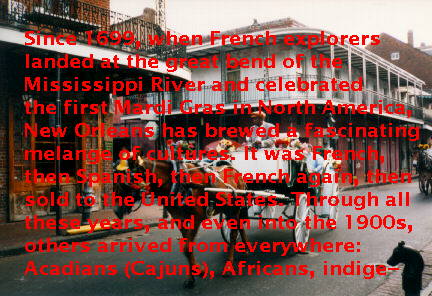}} \\
	
	\subfloat[IRCNN, PSNR=30.95dB]{%
		\includegraphics[height=2cm,width=3.5cm]{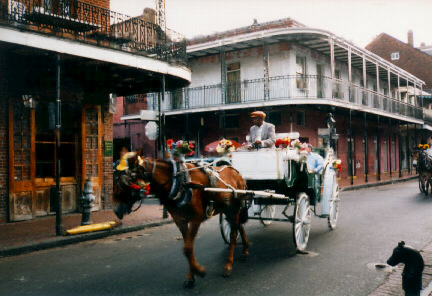}} \;
	\subfloat[FoE,PSNR=32.35dB]{%
		\includegraphics[height=2cm,width=3.5cm]{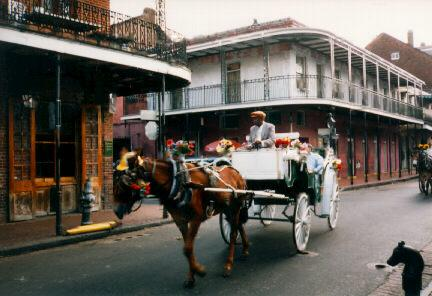}} \;
		\vspace*{-0.25cm}
	\caption{Comparison of superimposed text removal performance
	}\label{fig:textrem}
\end{figure}
\vspace*{-0.3cm}

We tested the performance of our algorithm of the classic image used in the original inpainting  paper\cite{Bertalmio:2000:II:344779.344972} for text removal. Quantitative evaluation on the data revealed that our model obtained a PSNR value of 30.95dB. For lack of blind inpainting methods, we compare our performance with non-blind inpainting method of Field-of-Experts(FoE) model\cite{1467533} and K-SVD model\cite{4392496}. For this image, FoE achieves PSNR value of 32.35dB while K-SVD(as reported in \cite{4392496}) achieves 32.45dB. We used the Matlab code provided by the authors, for evaluating the performance using FOE model. The time required by FoE for text removal was 584 seconds (using 24 ,$5\times 5 $ filters) while IRCNN took 5.6 seconds for the same task.The capability of our method for blind inpainting of complicated superimposed texts is a notable contribution of this paper.

\vspace*{-0.2cm}
\section{Conclusion}
\vspace*{-0.3cm}
We have presented a fully convolutional deep learning approach for image restoration of RGB images. The proposed approach learns an end-to-end mapping between noisy and clean image patches. Experimental evaluations on image denoising show that fully convolutional image denoising demonstrates competitive performance with the state-of-the-art methods. For image inpainting, our model solves the difficult problem of blind inpainting and successfully removes uniformly distributed impulse noise as well as sophisticated patterns like text with the impressive visual quality of reconstruction. These results show that proposed FCN model can indeed provide a good model for low-level image restoration problems. In addition to the demonstrated competitive accuracy, the proposed FCN based image restoration model is light-weight and feed-forward in structure which can be readily implemented in practical systems. 

\vspace*{-0.3cm}
\bibliographystyle{splncs}
\bibliography{egbib}

\end{document}